\def\year{2021}\relax
\documentclass[letterpaper]{article} 
\usepackage{aaai21}  
\usepackage{times}  
\usepackage{helvet} 
\usepackage{courier}  
\usepackage[hyphens]{url}  
\usepackage{graphicx} 
\usepackage{natbib}  
\usepackage{caption} 
\usepackage{array}
\usepackage{amssymb}
\usepackage{ragged2e}
\usepackage{bm}

\usepackage{multirow}
\usepackage{amsmath}

\usepackage{algorithmic}

\title{Communicative Message Passing for Inductive Relation Reasoning}
\author{Sijie Mai$^1$, Shuangjia Zheng$^1$, Yuedong Yang$^2$, Haifeng Hu$^2$\\
}
\affiliations{

Sun Yat-sen University\\

\{maisj, zhengshj9\}@mail2.sysu.edu.cn, \{yangyd25, huhaif\}@mail.sysu.edu.cn

}

\begin{document}
\renewcommand{\thefootnote}{\fnsymbol{footnote}}
\maketitle


\begin{abstract}
Relation prediction for knowledge graphs aims at predicting missing relationships between entities. Despite the importance of inductive relation prediction, most previous works are limited to a transductive setting and cannot process previously unseen entities. The recent proposed subgraph-based relation reasoning models provided alternatives to
predict links from the subgraph structure surrounding a candidate triplet inductively. However, we observe that these methods often neglect the directed nature of the extracted subgraph and weaken the role of relation information in the subgraph modeling. As a result, they fail to effectively handle the asymmetric/anti-symmetric triplets and produce insufficient embeddings for the target triplets. To this end, we introduce a \textbf{C}\textbf{o}mmunicative \textbf{M}essage \textbf{P}assing neural network for \textbf{I}nductive re\textbf{L}ation r\textbf{E}asoning, \textbf{CoMPILE}, that reasons over local directed subgraph structures and has a vigorous inductive bias to process entity-independent semantic relations. In contrast to existing models, CoMPILE strengthens the message interactions between edges and entitles through a communicative kernel and enables a sufficient flow of relation information. Moreover, we demonstrate that CoMPILE can naturally handle asymmetric/anti-symmetric relations without the need for explosively increasing the number of model parameters by extracting the directed enclosing subgraphs. Extensive experiments show substantial performance gains in comparison to state-of-the-art methods on commonly used benchmark datasets with variant inductive settings.

\end{abstract}

\section{Introduction}\label{sec:Introduction}
Knowledge graphs (KGs) are collections of factual information represented in the form of relational triplets. Each relation triplet can be organized as $(h, r, t)$ where $h$ and $t$ represents the head and tail entities and $r$ is the relation between $h$ and $t$. KGs have played a critical role across variety of tasks like Question Answering \cite{qa} , Semantic Search \cite{xiong2017explicit}, Dialogue Generation \cite{DG}  and many more. However, because of the limitations of human knowledge and extraction algorithms, they tend to suffer from incompleteness, that is, absent links in the KGs. Numerous methods have been developed to fulfill the gap between KGs and real-world knowledge, which are referred as link prediction or knowledge graph completion tasks. After condensing entities and relations into a low-dimensional vector space, these models predict missing facts by operating the entity and relation embeddings.

\footnote{$^1$These authors contributed equally to this work.}
\footnote{$^2$Corresponding authors.}

Despite the impressiveness of model performance, most of KG embeddings methods inherently assume a fixed set of entities in the graph and ignore the evolving nature of KGs. In fact, many real-world KGs are ever-evolving \cite{trivedi2017know}, with novel entities being added over time e.g., new users in social media platforms or new molecules in biomedical knowledge graphs. As these entities were unknown to the model during training, the model does not obtain their embeddings, and hence have no means to infer the relations for these entities. Thus, the ability to make predictions on new entities avoiding costly re-training from scratch is desirable for production-ready machine learning models.

Some efforts have been made to obtain the inductive embeddings for new entities using external resources \cite{wang2014knowledge,xie2016image,zhong2015aligning}. Although these approaches may be useful, they required extra computation over massive resources, which are time-consuming and may not always be feasible. An alternative is to induce probabilistic logic rules by enumerating statistical regularities and patterns embedded in the knowledge graph \cite{meilicke2018fine,yang2017differentiable}, which is entity-independent and hence inherently inductive. However, these methods suffer from scalability issues and are less expressive due to their rule-based nature.

Taking inspiration from the success of graph neural networks (GNNs) in graph data modeling, extensive studies applied GNNs to learn the embeddings for KGs \cite{RGCN,SACN,KBGT}, among which most of them were adopted in transductive settings. By extending a subgraph-based link prediction method that performs in the basic network \cite{zhang2018link}, recent work GraIL \cite{grail} shed light on the evolution issue for KG. The basic strategy behind GraIL can be split into three steps: (i) extracting the enclosing subgraph surrounding the target triplets, (ii) annotating the relative position of each entity in the extracted subgraph, and (iii) scoring the annotated subgraph with a GNN. It can be beneficial to the inference of entirely novel entities that not be surrounded by known nodes, and there is no need of the domain-related initial embedding of these emerging entities. The experimental results of applying GraIL demonstrate the feasibility of utilizing a subgraph-based model on inductive reasoning in the context of KGs.

One drawback of GraIL is that it ignores the directionality nature of the KG when extracting the enclosing subgraph for the target triplet. Although the direction is nominally preserved during message passing by aggregating the incoming information for an entity, the subsequent symmetric scoring function make it incapable of effectively handling binary relationships in KGs, especially for asymmetric (e.g.,
nominatedfor) and anti-symmetric relations (e.g., filiation). Moreover, the adopted vertex-based message passing network  \cite{xu2018powerful} weakens the role of relation embeddings, which violates the nature of the inductive relation reasoning because inductive setting is entity-independent and relies on the relation information to conduct inference.

Based on the above observations, we propose a novel Communicative Message Passing neural networks for Inductive reLation rEasoning (CoMPILE). In CoMPILE, we first extract directed enclosing subgraph for each triplet instead of an undirected one. A communicative message passing network framework is then extended to strengthen the information interactions between entities and relations while update both the edge and entity embeddings simultaneously. We also apply an edge-aware attention mechanism to aggregate the local neighborhood features and gather the global entity information to enrich the entity/relation representation. In contrast to previous efforts \cite{nickel2015holographic} that require an
explosion of the number of parameters to handle binary relations, our model is naturally capable of dealing with asymmetric and anti-symmetric relations by communicative message passing within directed subgraphs, without the unnecessary explosive increase of the number of parameters.

In brief, the main contributions are listed below:
\begin{itemize}
  \item Introducing a competitive inductive knowledge graph embedding model, CoMPILE, that fits the directional nature of knowledge graph and can naturally deal with asymmetric and anti-symmetric relations.
  \item Introducing a novel node-edge communicative message passing mechanism to strengthen the role of relation information in the subgraph and fit the nature of the inductive setting.
  \item Evaluating CoMPILE and several previously proposed models on three inductive datasets: our model achieves state-of-the-art AUC-PR and Hits@10 across most of them. We also extract new inductive datasets by filtering out the triplets that have no enclosing subgraph to evaluate the inductive relation reasoning more accurately.
\end{itemize}

\section{Related Work}\label{sec:Related}

\subsection{Transductive Relation Prediction}
Representation learning on KGs has been an active research area due to the wide applications of the resultant entity and relation embeddings. Typical KG embedding models include TransE \cite{TransE}, Distmult \cite{dismult}, ComplEx \cite{trouillon2017knowledge}, ConvE \cite{convE}, to name a few. While these kinds of methods process each triplet independently without the consideration of semantic and structure information embedded in rich neighborhoods. Recently, GNNs have been used to capture global structural information inherently stored in KGs and have been shown to achieve state-of-the-art performance on a variety of datasets \cite{RGCN,SACN,KBGT}. However, the above approaches on KG embedding mostly work in transductive settings.

\begin{figure}
\centering
\includegraphics[scale=0.58]{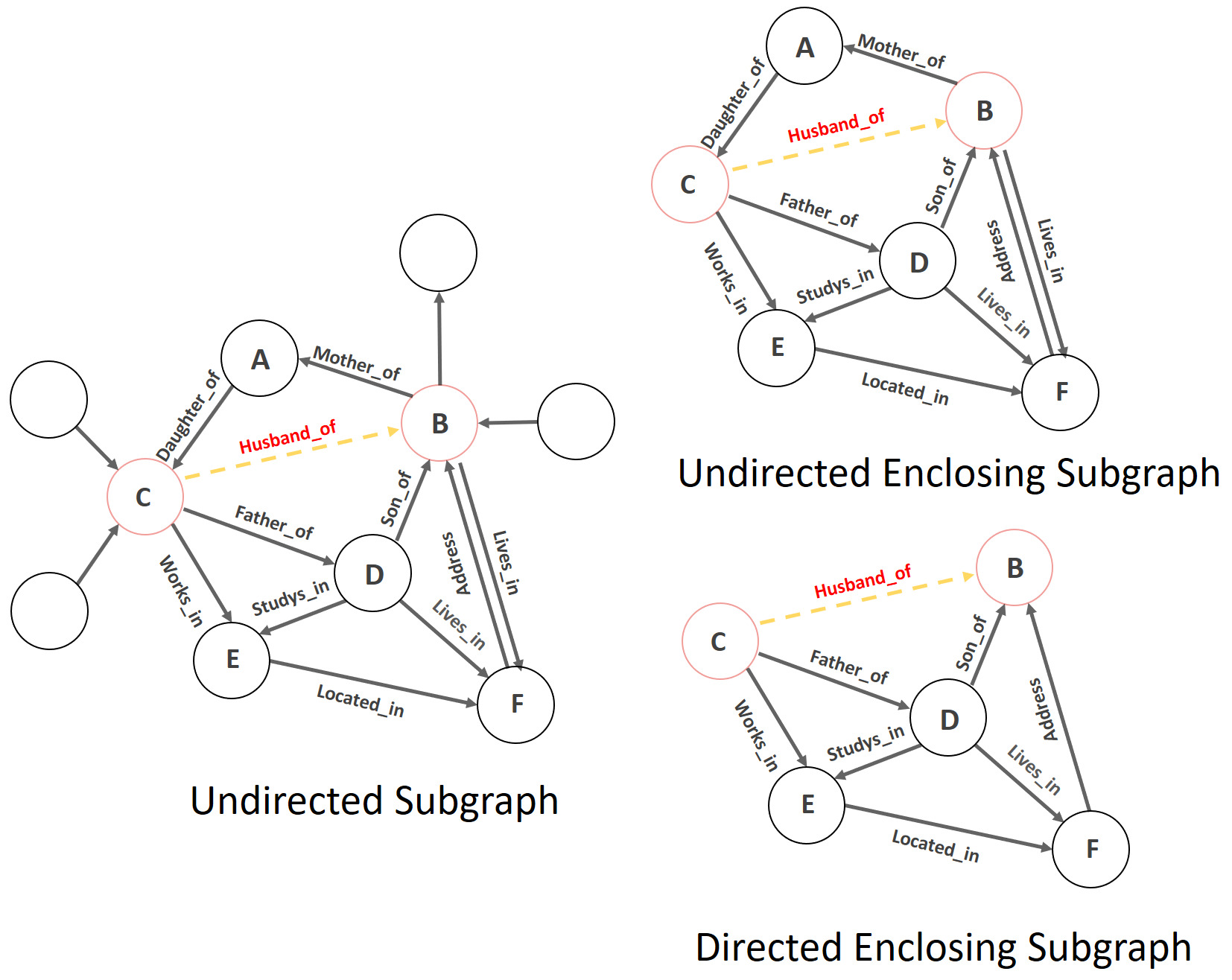}
\caption{\label{2}Comparisons of different subgraph extraction strategies. The left one is the undirected subgraph that considers all the $h$-hop neighbors of both the target head and tail. The subgraph on the right refer to the enclosing subgraph that only considers the $h$-hop common neighbors of target head and target tail.
}
\end{figure}

\subsection{Inductive Relation Prediction}
One research line of inductive representation learning is to introduce additional attributes such as description text or images to embed unseen entities \cite{xie2016image,xie2016representation,shi2017open}. Although the resultant embeddings can be utilized for KG completion, these methods heavily relied on the presence of external resources which is not present in many KGs. To relieve this issue, several inductive KG embedding models \cite{hamaguchi2017knowledge,wang2019logic} are proposed to aggregate the information of existing neighbors of an emerging entity with graph neural networks. However, both of these approaches demand the new nodes to be surrounded by known nodes and cannot handle entirely new graphs.

Another research line is rule-based approaches that use observed co-occurrences of frequent patterns in the knowledge graph to recognize logical rules \cite{galarraga2015fast}. They are inherently inductive since the logical-rules are independent of entities, but these approaches suffer from scalability issues and lack expressive power due to their rule-based nature. Inspired by these statistical rule-induction approaches, several differentiable rule learners including NeuralLP \cite{yang2017differentiable}, RuleN \cite{meilicke2018fine}, and DRUM \cite{DRUM} are proposed to learn the logical rules as well as confidence scores from KGs in an end-to-end paradigm. However, they did not take account of the neighbor structure surrounding the predicted relations, hence is not expressive enough when the paths between head and tail entities are sparse. This set of methods, together with GraIL \cite{grail} constitute our baselines.

\section{CoMPILE}

\subsection{Denotations and Task Definition}
A target triplet in the knowledge graph is denoted as $(s, r, t)$ where $s$, $r$, and $t$ refers to the head entity, relation, and tail entity, respectively. Inductive relation reasoning aims to score the plausibility of target triplet $(h_T, r_T, t_T)$ , where the representations of $h_T$ and $t_T$ are not available during prediction. In this work, we use an enclosing directed subgraph to represent the target triplet $(h_T, r_T, t_T)$. The enclosing subgraph between the target head and tail is denoted as $G=(V, E)$ where $V$ and $E\in V\times V$ denotes the set of nodes and observed edges in the subgraph $G$, respectively. We use $N_e$ to represent the number of edges in the subgraph. The embeddings of nodes is denoted as
$\bm{N}\in \mathbb{R}^{N_n\times d_n}$ where $N_n$ is the number of the nodes in the subgraph.  The relation embedding is denoted as $\bm{R}\in \mathbb{R}^{N_r\times d}$ ($N_r$ is the number of relations), which is parameterized as a learnable matrix updated by gradient descent and is shared across train and test graphs. We define the head-to-edge, relation-to-edge, and tail-to-edge adjacency matrix as $\bm{A^{he}}\in \mathbb{R}^{N_n\times N_e}$, $\bm{A^{re}}\in \mathbb{R}^{N_r\times N_e}$, and $\bm{A^{te}}\in \mathbb{R}^{N_n\times N_e}$, which maps the tail, relation, and head to the corresponding edge, respectively. The values in the adjacency matrices are either 0 or 1 where 0 denotes no connections.

\subsection{Directed Subgraph Extraction}
In this section, we illustrate the procedure to extract the directed enclosing subgraph. Different from GraIL \cite{grail} that extracts undirected enclosing subgraph between the target head and tail which ignores the direction of the target triplet, inspired by the mechanism that human uses to infer logical rules, we demonstrate the superiority of directed enclosing subgraph. A triplet has direction, representing the relation from head to tail. Considering two triplets $(h_T, r_T, t_T)$ and $(t_T, r_T, h_T)$, if we use undirected enclosing subgraph, then the predictions are likely to be very close for these two triplets because the enclosing subgraphs are the same. However, only one of them is true if the relation $r_T$ is asymmetric. Therefore, we need to use directed enclosing subgraph to handle these kinds of relation more effectively. Moreover, by this means, we can solve the direction problem in KGs without increasing the model complexity (the time complexity even decreases since the directed enclosing subgraph is obtained by pruning the undirected enclosing subgraph, as shown in Fig.~\ref{2}).

The enclosing subgraph is restricted by the hop number $h$, and $h+1$ is the maximum distance from target head to target tail. To extract the $h$-hop directed enclosing subgraph, we first introduce the incoming neighbors and outgoing neighbors. Given a triplet $(s, r, t)$, we define $s$ as the 1-hop incoming neighbor of $t$, and $t$ as the 1-hop outgoing neighbor of $s$, and $h$-hop incoming/outgoing neighbors vice verse.
Firstly, we extract $h$-hop outgoing neighbors of target head and the $h$-hop incoming neighbors of target tail. Then we extract the 1-hop incoming neighbor of target tail. If the $h$-hop outgoing neighbors of target head and the 1-hop incoming neighbor of target tail have common entities, then the directed subgraph between target head and tail is existed. Afterwards, We find the common entities (nodes) of $h$-hop outgoing neighbors of target head and the $h$-hop incoming neighbor of target tail (if target head or tail are not in the common entities, we need to add them into common entities). Finally we add the edges (triplets) whose head and tail both belong to the common entities to construct the subgraph. By this means, the  maximum distance from target head to tail would become $h+1$ even though we only extract $h$-hop neighbors of them.

\subsection{Node/Edge Embedding Initialization}

Since the entities are unseen, we need to define an entity-independent embedding for each entity (node) in the subgraph. Similar to GraIL \cite{grail}, we initialize the node embedding by the distances to the target head and target tail to capture the relative position of each node in the subgraph. The node embedding for node $i$ is defined as $\bm{N_i}=\text{one-hot}(d_{hi}) \oplus \text{one-hot}(d_{it})\in \mathbb{R}^{2(h+2)}$ where $d_{hi}$ denotes the minimum distance from the target head to node $i$, and $d_{it}$ denotes the minimum distance from the node $i$ to target tail (note that the distance definition here also meets the `directed' requirement). For edge $i$, i.e. $(h_i, r_i, t_i)$, the initialized edge embedding is defined as $\bm{E}_i=\bm{N}_{h_i}\oplus \bm{R}_{r_i}\oplus \bm{N}_{t_i}\in \mathbb{R}^{4(h+2)+d}$.

\subsection{Directed Subgraph Modeling}
\begin{figure}
\centering
\includegraphics[scale=0.8]{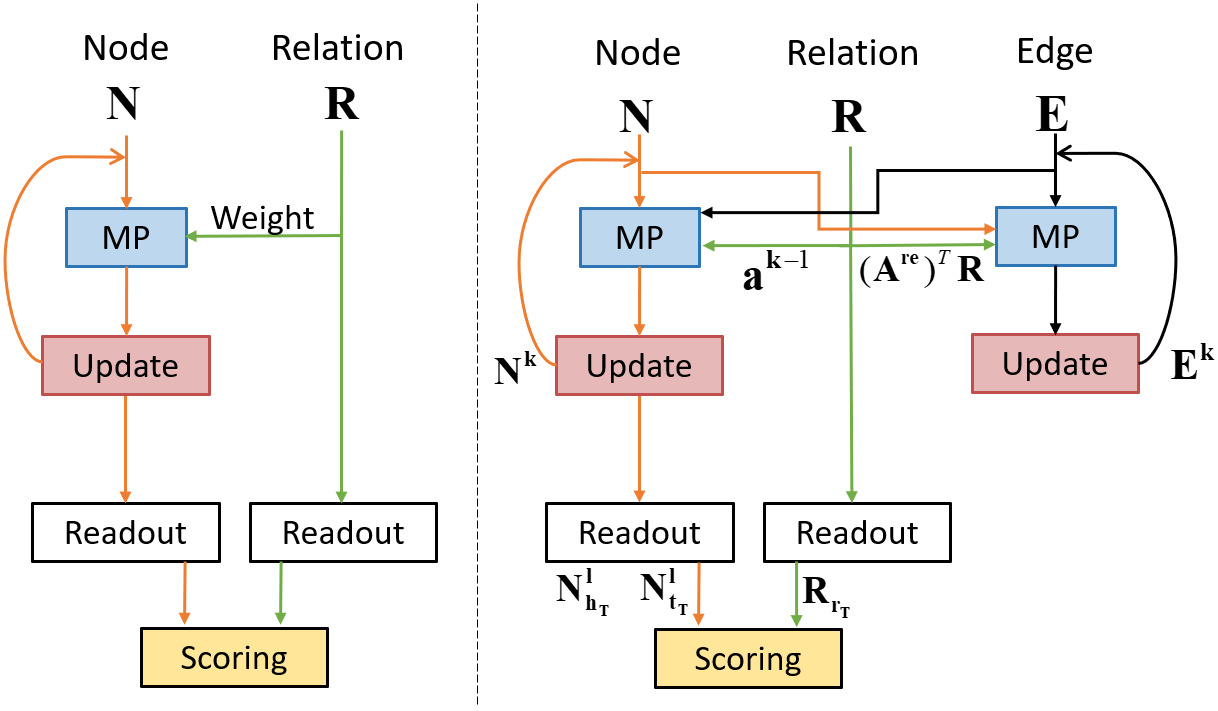}
\caption{\label{1}Comparison of Message Passing Mechanism in GraIL and Our Model. `MP' in the figure denotes `message passing', and the left and right figure refers to the message passing mechanism in GraIL and our model, respectively. Our method explicitly models edge embedding to strengthen the flow of the relation information and bidirectionally communicate the nodes and edges.
}
\end{figure}

The message passing model in GraIL \cite{grail} is a simple R-GCN \cite{RGCN} with edge attention, which ignores separately modeling edge embedding and ignores the bidirectional communication between edges and nodes. Moreover, GraIL uses a node-to-node message passing mechanism where the relation information is only used for computing the weights for the neighboring nodes. However, the inductive relation reasoning
should be entity-free (i.e., node-free in the subgraph case), where relation plays a dominant role while the entities cannot provide deterministic information during inference. Thus, the node-to-node message passing mechanism in Grail weakens the role of relations and violates the nature of the inductive knowledge graph. To this end, we design a brand new message passing architecture to model the inductive enclosing subgraph by iteratively  communicating and enhancing the edge and node embeddings (see Fig.~\ref{1}). Since the edge information is mainly provided by the corresponding relation, the nodes can learn to better aggregate the relation information in the subgraph during the node-edge interactions, such that the model can learn to infer the relations between the target head and tail based on the relation information presented in the subgraph.

Our node-edge interaction mechanism is inspired by CMPNN \cite{DMPNN}. Nevertheless, during the node-edge interactions of CMPNN, the representation of each edge is updated by the head node embedding and its inverse edge (which is not available in our task), but neglecting the tail node embedding and the relation embedding. In our framework, we introduce a new node-edge communication mechanism that considers both head, relation, and tail to update the edge embedding. We describe our message passing model in detail in the following sections.

Firstly, the node and edge representations are mapped to the same dimensionality $d$ using the following equations ($d$ is also the dimensionality of the relation embedding):
\begin{equation}
\label{eq0}
\setlength{\abovedisplayskip}{3pt}
\setlength{\belowdisplayskip}{3pt}
  \bm{N}^{0}= f_1(\bm{N}\bm{W}_{n}^0),\ \bm{E}^{0}= f_1(\bm{E}\bm{W}_{e}^0)
\end{equation}
where $f_1$ denotes the nonlinear activation function to increase the nonlinear expressive power of the model, $\bm{W}_{n}^0 \in \mathbb{R}^{2(h+2)\times d}$ and $\bm{W}_{e}^0  \in \mathbb{R}^{(4(h+2)+d)\times d}$ are the learnable parametric matrix, $\bm{N}^{0} \in \mathbb{R}^{N_n\times d}$ and $\bm{E}^{0} \in \mathbb{R}^{N_e\times d}$ is the transformed node and edge embedding, respectively.

\textbf{Node Embedding Updating}:
The node embedding is updated for totally $l$ iterations. At each iteration (we take iteration $k$ as an example in the following equations), edge embedding is required for updating the node embedding in our node-edge interaction mechanism. Firstly, to highlight the edges that are highly related to the target triplet, we design an \textbf{enhanced edge attention}. In the edge attention of GraIL \cite{grail}, only target relation is utilized for predicting the importance of edges. In contrast, we utilize all the target head, target relation, and target tail to highlight the edges that have a close connection to the target triplet, which is more comprehensive. We use the whole triplet to conduct attention because the nodes can aggregate the relation information during node-edge interactions and thereby the updated node embeddings are also informative.  For edge $i$ ($ 1\leq i\leq N_e$), the equations for the enhanced edge attention are presented as follows:
\begin{equation}
\label{eq1}
\setlength{\abovedisplayskip}{3pt}
\setlength{\belowdisplayskip}{3pt}
  \bm{a}_{i}^{k-1}=(\bm{N}_{h_i}^{k-1} + \bm{R}_{r_i} - \bm{N}_{t_i}^{k-1}) \oplus (\bm{N}_{h_T}^{k-1} + \bm{R}_{r_T} - \bm{N}_{t_T}^{k-1})
\end{equation}
\begin{equation}
\label{eq10}
\setlength{\abovedisplayskip}{3pt}
\setlength{\belowdisplayskip}{3pt}
  a_{i}^{k-1}= \sigma(f_1(\bm{a}_{i}^{k-1}\bm{W}_{a_1}^{k-1})\bm{W}_{a_2}^{k-1})
\end{equation}
\begin{equation}
\label{eq11}
\setlength{\abovedisplayskip}{3pt}
\setlength{\belowdisplayskip}{3pt}
  \bm{E}_i^{(k-1)_a} = a_{i}^{k-1} \bm{E}_i^{k-1}
\end{equation}
where $\bm{E}_i^{(k-1)_a}$ is the attentive embedding for edge $i$, $a_{i}^{k-1}$ is a scalar that represents the weight for edge $i$, $\oplus$ denotes feature concatenation,  $\bm{N}_{h_T}^{k-1} + \bm{R}_{r_T} - \bm{N}_{t_T}^{k-1}$ denotes the embedding for target triplet, and we use $\bm{N}_{h_i}^{k-1} + \bm{R}_{r_i} - \bm{N}_{t_i}^{k-1}$ to denote edge $i$ so as to be consistent with the representation of the target triplet.

Then we use the attentive edge embedding to update the node representation:
\begin{equation}
\label{eq12}
\setlength{\abovedisplayskip}{3pt}
\setlength{\belowdisplayskip}{3pt}
  \bm{N}^{k}_{agg} = \bm{A^{te}}  \bm{E}^{(k-1)_a}
\end{equation}
\begin{equation}
\label{eq13}
\setlength{\abovedisplayskip}{3pt}
\setlength{\belowdisplayskip}{3pt}
  \bm{N}^{k} = f_1((\bm{N}^{k}_{agg} + \bm{N}^{k-1})\bm{W}^{k}_n)
\end{equation}
where $\bm{N}^{k}_{agg}$ denotes the node aggregation information, $\bm{W}^{k}_n\in \mathbb{R}^{d\times d}$ represents the parametric matrix for node embedding at iteration $k$, $\bm{E}^{(k-1)_a}$ is the attentive edge embedding, $\bm{A^{te}}$ is the tail-to-edge adjacency matrix that connects each edge to its tail. Note that to preserve the directed nature of our model, for an edge $(h_i, r_i, t_i)$, its embedding is only used to update the tail $t_i$ but not $h_i$. Coupling with the directed subgraph, it ensures that the information only flows from target head to target tail and the reverse flow is forbidden. By using the edge embedding to update node embedding, the target tail embedding can aggregate all the relations along with their relative positions occurred from the target head to target tail in the subgraph (the relative positions are provided by the node embedding), which provides powerful relational inference ability.

In the last iteration of the node embedding updating, similar to CMPNN \cite{DMPNN}, we use a multi-layer perception network followed by a Gated Recurrent Unit (GRU) \cite{Cho2014Learning} to replace Eq.~\ref{eq13} to increase the expressive power of the network, as shown in the following equations:
\begin{equation}
\label{eq14}
  \bm{N}^{l'} = \text{CommunicationMLP} (\bm{N}^{l}_{agg} \oplus \bm{N}^{l-1} \oplus \bm{N}^{0})
\end{equation}

\begin{equation}
\label{eq15}
      \bm{N}^{l} = \text{GRU}(\bm{N}^{l'})
\end{equation}
where CommunicationMLP is the multi-layer perception network that communicates the node aggregation information $\bm{N}^{l}_{agg}$, node embedding $\bm{N}^{l-1}$, and the original transformed node embedding $\bm{N}^{0}$ (we add $\bm{N}^{0}$ to perform residual learning \cite{4}). Note that GRUs requires the inputs to be ordered, while our node embeddings are inherently partly ordered since the  nodes are arranged according to the distance to target head in the ascending order.

\textbf{Edge Embedding Updating}:
The edge embedding is updated for totally $l-1$ iterations. To update the edge embedding, node embedding is required in our node-edge interaction mechanism. We define inverse mappings from node to edge and relation to edge, which are denoted as:
\begin{equation}
\label{eq2}
\setlength{\abovedisplayskip}{3pt}
\setlength{\belowdisplayskip}{3pt}
  \bm{E}_{agg}^k=(\bm{A^{he}})^T\bm{N}^{k} + (\bm{A^{re}})^T\bm{R} -  (\bm{A^{te}})^T\bm{N}^{k}
\end{equation}
where $T$ denotes matrix transpose,  $(\bm{A^{he}})^T\bm{N}^{k}$ aggregates the head information to edge,  $(\bm{A^{re}})^T\bm{R}$ aggregates the relation information to edge, and  $(\bm{A^{te}})^T\bm{N}^{k}$ aggregates the tail information to edge. The definition of edge aggregation information $\bm{E}_{agg}^k$ in Eq.~\ref{eq2} meets the directed requirement and is consistent across the model. Then we use the aggregation information  to update edge representation:
\begin{equation}
\label{eq3}
\setlength{\abovedisplayskip}{3pt}
\setlength{\belowdisplayskip}{3pt}
  \bm{E}^{k'}= f_1( \bm{E}^{k-1} + f_2(\bm{E}_{agg}^k))
\end{equation}

\begin{equation}
\label{eq4}
\setlength{\abovedisplayskip}{3pt}
\setlength{\belowdisplayskip}{3pt}
  \bm{E}^{k}= f_1(\bm{E}^{k'}\bm{W}^{k}_e + \bm{E}^{0})
\end{equation}
where $f_1$ and $f_2$ denotes nonlinear activation function to increase the nonlinear modeling capacity of the model.  We add the $\bm{E}^{0}$ to update the edge embedding in Eq.~\ref{eq4} to perform residual learning. Edge dropout is also performed on $\bm{E}^{k}$.

\textbf{Scoring Function Definition}:
GraIL designed an asymmetric scoring function for subgraph inductive learning by concatenating four related vectors:

\begin{equation}
\label{eq5}
\setlength{\abovedisplayskip}{3pt}
\setlength{\belowdisplayskip}{3pt}
  \bm{S} = \bm{W}(\bm{h}_{G}^l \oplus \bm{h}_{h_T}^l \oplus \bm{h}_{t_T}^l \oplus \bm{e}_{r_T})
\end{equation}
where $\bm{h}_{G}^l$ denotes the subgraph representation, $\bm{h}_{h_T}^l$ and $\bm{h}_{t_T}^l$ denote the hidden vectors of head and tail entities, $\bm{e}_{r_T}$ is a learned embedding of the target relation. This scoring function is symmetric as both the relation embedding and subgraph embedding are undirected. To alleviate this problem, we adopt the idea of TransE \cite{TransE} to design the scoring function so as to preserve directed nature of our model as well as to be consistent with the definition of the edge information. The scoring function is defined as:
\begin{equation}
\label{eq5}
\setlength{\abovedisplayskip}{3pt}
\setlength{\belowdisplayskip}{3pt}
  \bm{S} = f_2(\bm{N}_{h_T}^l + \bm{R}_{r_T} - \bm{N}_{t_T}^l )
\end{equation}
where $\bm{N}_{h_T}^l$, $\bm{R}_{r_T}^l$, and $\bm{N}_{t_T}^l$ denotes the final representation of target head, target relation, and target tail, respectively.
We than use a two-layer fully-connected network on $\bm{S}$ to infer the score of the target triplet $(h_T, r_T, t_T)$.

\section{Experiments}\label{sec:Experiments}
In our experiments, we aim to answer the following questions: 1) Does our CoMPILE outperforms state-of-the-art methods on the commonly used datasets? 2) Does our proposed message passing network better than the RGCN + edge attention in GraIL? 3) Does the directed subgraph outperform the undirected one? 4) What are the importance of the components in CoMPILE? 5) Can CoMPILE deals with asymmetric relations better than the other methods?

\subsection{Datasets}
WN18RR \cite{convE}, FB15k-237 \cite{FB15k-237}, and Nell-995 \cite{NELL-995} are commonly used datasets that are originally developed for transductive relation prediction. Teru et al. \cite{grail} extracts four versions of inductive datasets for each dataset. 
Each inductive dataset constitutes of train and test graphs, where the test graph contains entities that are not presented in train graph. The subgraph for each triplet is extracted from the train or test graph.

\textbf{Our Post-processed Datasets}: In our experiment, we found that the inductive datasets constructed by GraIL \cite{grail} have many empty-subgraph triplets  (no valid edge exists in the enclosing subgraph of target head and  tail under hop $h$), especially for WN18RR. Moreover, since in  GraIL, the negative triplets are randomly sampled, leading to a significant number of  empty-subgraphs in negative triplets. This results in inaccurate evaluation of the performance, for the reason that under the subgraph reasoning structure, it is almost impossible to infer the relation between two entities if there are no valid edges in the subgraph (these triplets also cannot work in rule-based structures such as RuleN \cite{meilicke2018fine}).

Therefore, we construct new inductive datasets by filtering out the triplets in the original inductive datasets that have no subgraph under hop $h$ and constructing negative triplets that have subgraphs between the fake target head and fake target tail. Specifically, we extract three versions of inductive datasets each for FB15k-237 and NELL-995 datasets. Since there are no enough non-empty negative triplets for each positive triplet to perform Hits@10 experiment in the original inductive WN18RR datasets, we only extract the inductive datasets for NELL-995 and FB15k-237. The statistics of the datasets is shown in Appendix. Note that the subgraph here refers to undirected enclosing subgraph extracted in GraIL. An undirected subgraph is existed for two entities does not necessarily means that a directed subgraph is existed. If a directed subgraph does not existed, we will extract the undirected subgraph for these two entities.

\subsection{Experimental Details}
To be consistent with the prior methods, we use AUC-PR and Hits@10 to evaluate the models. Similar to GraIL \cite{grail}, to compute AUC-PR, we sample one negative triplet for each test triplet and evaluate which triplet has larger score. For Hits@10, we compare the true triplet with the sampled negative triplets in terms of the scores, to see whether the true triplet can rank the top 10. The negative triplets are obtained by replacing the head or tail of the test triplets with other entities. For the original inductive datasets, the negative triplets are randomly sampled and do not consider whether they have enclosing subgraph. For our extracted inductive datasets, we ensure that the negative triplets can also have an enclosing subgraph.

We implement our model on Pytorch. We use Adam \cite{Kingma2014Adam} as optimizer with learning rate being 0.001. The hop number $h$ is set to 3 which is consistent with GraIL. We train the model for four times and average the testing results to obtain the final performance. The number of iterations $l$ is set to 3. For more details, please refer to our codes at: \url{https://github.com/TmacMai/CoMPILE_Inductive_Knowledge_Graph}.

We evaluate our model with the following baselines: Neural-LP \cite{yang2017differentiable}, DRUM  \cite{DRUM}, RuleN \cite{meilicke2018fine}, and GraIL \cite{grail}. For the detailed introduction of these methods, please refer to the introduction and  related work sections.

\begin{table*}[!htb]
\centering
\resizebox{1.92\columnwidth}{!}{\begin{tabular}{c|c|c|c|c|c|c|c|c|c|c|c|c}
 \hline
  & \multicolumn{4}{c|}{WN18RR} &  \multicolumn{4}{c|}{FB15k-237} &  \multicolumn{4}{c}{NELL-995} \\
 \hline
 Methods  & v1  & v2 & v3 & v4 & v1 & v2 & v3 & v4 & v1 & v2 & v3 & v4 \\
 \hline
 Neural-LP & 86.02 & 83.78 & 62.90 & 82.06 & 69.64 & 76.55 & 73.95 & 75.74 & 64.66 & 83.61 & 87.58 & 85.69 \\
 DRUM & 86.02 & 84.05 & 63.20 & 82.06 & 69.71 & 76.44 & 74.03 & 76.20 & 59.86 & 83.99 & 87.71 &  85.94\\
 RuleN & 90.26 & 89.01 & 76.46 & 85.75 & 75.24 & 88.70 & 91.24 & 91.79 & 84.99 & 88.40 & 87.20 & 80.52\\
 GraIL &  94.32 &  94.18 &  85.80 &  92.72 &  84.69 &  90.57 &  91.68 &  94.46 & \textbf{86.05} &  92.62 &  93.34 & \textbf{87.50}\\
 \hline
    CoMPILE & \textbf{98.23} & \textbf{99.56} & \textbf{93.60} & \textbf{99.80} & \textbf{85.50} & \textbf{91.68} & \textbf{93.12} & \textbf{94.90} & 80.16 & \textbf{95.88} & \textbf{96.08} & 85.48\\
 \hline
 \end{tabular}}
  \caption{ \label{t1}Compared with Baselines on the Original Inductive Datasets (AUC-PR). The best performance is highlighted.}
\end{table*}%

\begin{table*}[!htb]
\centering
\resizebox{1.92\columnwidth}{!}{\begin{tabular}{c|c|c|c|c|c|c|c|c|c|c|c|c}
 \hline
  & \multicolumn{4}{c|}{WN18RR} &  \multicolumn{4}{c|}{FB15k-237} &  \multicolumn{4}{c}{NELL-995} \\
 \hline
 Methods  & v1  & v2 & v3 & v4 & v1 & v2 & v3 & v4 & v1 & v2 & v3 & v4 \\
 \hline
 Neural-LP & 74.37 & 68.93  & 46.18 & 67.13 & 52.92 & 58.94 & 52.90 & 55.88 & 40.78 & 78.73 & 82.71 & \textbf{80.58} \\
 DRUM & 74.37 & 68.93 & 46.18 & 67.13 & 52.92 & 58.73 & 52.90 & 55.88 & 19.42 & 78.55 & 82.71 & \textbf{80.58}\\
 RuleN & 80.85 & 78.23 & 53.39 & 71.59 & 49.76 & 77.82 & \textbf{87.69} & 85.60 & 53.50 & 81.75 & 77.26 & 61.35\\
 GraIL & 82.45 & 78.68 & 58.43 & 73.41 & 64.15 & 81.80 & 82.83 & \textbf{89.29} & \textbf{59.50} & 93.25 & 91.41 & 73.19\\
 \hline
  CoMPILE& \textbf{83.60} & \textbf{79.82} & \textbf{60.69} & \textbf{75.49} & \textbf{67.64} & \textbf{82.98} & 84.67 & 87.44 & 58.38 & \textbf{93.87} & \textbf{92.77} & 75.19\\
 \hline
 \end{tabular}}
  \caption{ \label{t2}Compared with Baselines on the Original Inductive Datasets (Hits@10).}
\end{table*}%

\subsection{Results and Discussions}
\subsubsection{Compare with Baselines on Original Inductive Datasets}\label{sec:one}
In this section, we compare our proposed message passing network with other baselines on the original inductive datasets proposed by GraIL. Note that this experiment is performed to evaluate the effectiveness of our node-edge message passing network, and the other settings such as the enclosing subgraph, loss function, and node embedding initialization are the same as those in GraIL. As presented in Table~\ref{t1} and Table~\ref{t2}, our CoMPILE demonstrates improvement on the majority of the original inductive datasets in terms of both the AUC-PR and Hits@10 evaluation metrics. Specifically, the improvement is significant on the AUC-PR for WN18RR inductive datasets, on which our CoMPILE outperforms SOTA method GraIL by a large margin. The results indicate that the proposed message passing network
shows marked improvement over the R-GCN with attention in GraIL, which highlights the necessity of the bidirectional communication between the node and edge embeddings as well as the effectiveness of strengthening the role of relation information in the subgraph modeling. 

\begin{table*}[!htb]
\centering
\resizebox{2.1\columnwidth}{!}{\begin{tabular}{c|c|c|c|c|c|c|c|c|c|c|c|c}
 \hline
  &   \multicolumn{6}{c|}{FB15k-237} &  \multicolumn{6}{c}{NELL-995} \\
 \hline
   & \multicolumn{2}{c|}{v1} & \multicolumn{2}{c|}{v2} & \multicolumn{2}{c|}{v3} &  \multicolumn{2}{c|}{v1} & \multicolumn{2}{c|}{v2} & \multicolumn{2}{c}{v3} \\
    \hline

    & AUC-PR & Hits@10  & AUC-PR & Hits@10   & AUC-PR & Hits@10   & AUC-PR & Hits@10   & AUC-PR & Hits@10     & AUC-PR & Hits@10\\

 \hline
 GraIL & \textbf{78.84} & 61.25 & 80.76 & 64.93 & 82.25 & 68.13   & \textbf{68.47} & 58.63 & 84.07 & \textbf{76.52} & 81.42 &  75.12 \\
  CoMPILE& 78.06 & \textbf{63.96} & \textbf{81.09} & \textbf{68.17} & \textbf{83.21} & \textbf{71.67} & 68.13 & \textbf{61.38} & \textbf{85.67} & 76.11 & \textbf{82.88} & \textbf{76.83}\\
  \hline
 \end{tabular}}
  \caption{ \label{t3}Compared with Baselines on Our Extracted Inductive Datasets. }
\end{table*}%

\subsubsection{Compare with Baseline on Our Inductive Datasets}
As demonstrated in the Dataset section, the original inductive dataset contains empty-subgraph triplets under hop $h$ which leads to inaccurate evaluation of the performance. Therefore, we further evaluate CoMPILE on our post-processed inductive datasets that have filtered out the empty-subgraph triplets. For comparison, we implement and evaluate the R-GCN with edge attention which is the message passing method in GraIL, and the results are presented in Table~\ref{t3}. Note that the enclosing subgraphs are directed in both the models. As shown in Table~\ref{t3}, CoMPILE still performs better than the GraIL on the majority of evaluation metrics in our extracted inductive datasets, which further suggests the superiority of the message passing method in CoMPILE. For \textbf{model complexity}, the results suggest that our CoMPLIE has a total number of 35,969 parameters, while the number of parameters of GraIL is 29,264. CoMPILE has slightly more parameters than GraIL, mainly because we model both the edge and node embeddings to enhance the flow of relation information in our CoMPILE.

\begin{table*}[!htb]
\centering
\resizebox{2.1\columnwidth}{!}{\begin{tabular}{c|c|c|c|c|c|c|c|c|c|c|c|c}
 \hline
  &   \multicolumn{6}{c|}{FB15k-237} &  \multicolumn{6}{c}{NELL-995} \\
 \hline
   & \multicolumn{2}{c|}{v1} & \multicolumn{2}{c|}{v2} & \multicolumn{2}{c|}{v3}  & \multicolumn{2}{c|}{v1} & \multicolumn{2}{c|}{v2} & \multicolumn{2}{c}{v3} \\
    \hline
    & AUC-PR & Hits@10  & AUC-PR & Hits@10   & AUC-PR & Hits@10   & AUC-PR & Hits@10   & AUC-PR & Hits@10      & AUC-PR & Hits@10\\
 \hline
 Undirected & 77.80 & 59.79 & 80.89 & 65.98 & 81.45 & 68.82 & 63.25 & 58.88 & 79.33 & 68.92 & \textbf{83.73} & 76.66\\
  Directed & \textbf{78.06} & \textbf{63.96} & \textbf{81.09} & \textbf{68.17} & \textbf{83.21} & \textbf{71.67}  & \textbf{68.13} & \textbf{61.38} & \textbf{85.67} & \textbf{76.11} & 82.88 & \textbf{76.83}\\
  \hline
 \end{tabular}}
  \caption{ \label{t4}Compared between the Undirected and Directed Subgraphs on CoMPILE. }
\end{table*}%

\subsubsection{Directed Subgraph vs. Undirected Subgraph}
In this section, we investigate whether the performance has improved when the undirected enclosing subgraph is replaced by the directed one. From the results in Table~\ref{t4} we can infer that the version of CoMPILE with directed subgraph significantly outperforms the version of  CoMPILE with undirected subgraph almost in all the inductive datasets and evaluation metrics, demonstrating the necessity of effectively handling the direction problem in knowledge graph. This is in line with our expectations as both the FB15k-237 and NELL-995 contain a significant number of asymmetric and anti-symmetric relations \cite{wang2018evaluating}. It is also worth mentioning that dealing with the directed subgraph is more efficient in terms of the \textbf{time complexity}, for the reason that the directed subgraph can be viewed as the cropped version of the undirected one. Although being more efficient, our directed subgraph still outperforms undirected subgraph in terms of AUC-PR and Hits@10 metrics, indicting that a directed subgraph is a better choice to infer the relation between two unseen entities.

\begin{table}[!htb]
\centering
\resizebox{.98\columnwidth}{!}{\begin{tabular}{c|c|c|c|c}
 \hline
    & \multicolumn{2}{c|}{v1} & \multicolumn{2}{c}{v2}\\
 \hline
  & AUC-PR & Hits@10  & AUC-PR & Hits@10\\
  \hline
W/O EEA & 75.79 & 61.35 & 79.26 & 67.16\\
Edge Attention & 77.07 & 61.04 & 78.53 & 66.46\\
\hline

W/O EEU & 76.69 & 60.63 & 79.64 & 62.44\\
W/O Relation in EEU & 77.74 & 60.73 & 79.25 & 62.60 \\
 \hline
 CoMPILE & \textbf{78.06} & \textbf{63.96} & \textbf{81.09} & \textbf{68.17}\\
 \hline
 \end{tabular}}
  \caption{ \label{t5}Ablation Studies on Inductive FB15k-237 datasets. The `EEA' and `EEU' refers to Enhanced Edge Attention and Edge Embedding Updating, respectively. }
\end{table}%

\subsubsection{Ablation Studies}
We perform ablation studies to investigate the  contributions of different components in our architecture. For the enhanced edge attention, we remove it from our model to analyze its contribution (see the case of `W/O EEA' in Table~\ref{t5}). Moreover, we compare our enhanced edge attention with the edge attention in GraIL which only uses the target relation to conduct attention (see the case of `Edge Attention').
Furthermore, we investigate whether the updating of the edge embedding gives rise to the overall performance (see the case of `W/O EEU'). Additionally, we also remove the relation information in the edge embedding updating to analyze the importance of relation information (see the case of `W/O Relation in EEU', and Eq.~\ref{eq2} will become $\bm{E}_{agg}^k=(\bm{A^{he}})^T\bm{N}^{k} -  (\bm{A^{te}})^T\bm{N}^{k}$ in such circumstance).

From Table~\ref{t5} we find that CoMPILE significantly outperforms the version of CoMPILE whose enhanced edge attention is removed, demonstrating the effectiveness of our enhanced edge attention. Specifically, enhanced edge attention yields over 2\% improvement on both AUC-PR and Hits@10 metrics of the inductive FB15k-237-v1 dataset. In contrast, the improvement of the edge attention in GraIL is relatively small, and even the model without attention outperforms the model with edge attention on inductive FB15k-237-v2 dataset. These results suggest that it is of great significance to consider all the information in target triplet to determine which edges are considered to be important.

From the result of `W/O EEU' in Table~\ref{t5}, we can notice that the performance drops significantly on all the evaluation metrics when the edge embedding updating component is removed. Compared to completely removing the edge embedding updating module, removing the relation information in the edge embedding updating only yields slight improvement and is still clearly weaker than the case that the relation information is presented. These results demonstrate that the edge embedding updating module is a critical component that contributes to the marked improvement of our CoMPILE, and the relation information plays a very important role in the edge embedding updating. It further proves our claim that the relation is critical in inductive setting and highlights the necessity of strengthening the flow of relation information in the subgraph modeling.


\subsubsection{Case Study for Asymmetric Relations}
To demonstrate that our model can naturally handle asymmetric/anti-symmetric relations to some extent (by using the specific scoring function and edge definition, etc) even provided with the undirected enclosing subgraph, we select five relations that are asymmetric to evaluate our CoMPILE and GraIL. 
We use two negative triplets sampling strategies, where the first one is the standard operation that replaces the head or tail of the true triplets with other entities, and the second one is to exchange the head and tail of the test triplets. The results are presented in Table~\ref{t6}. Interestingly, we observe that the AUC scores of CoMPILE have no significant difference in this two negative triplets sampling strategies, indicating that CoMPILE can effectively
distinguish the false triplet $(t, r, h)$ from the true triplet $(h, r, t)$. In contrast, the performance of GraIL drops significantly when the sample strategy changes to the second one, which suggests that it cannot well distinguish these two triplets, mainly for the reason that the GraIL does not well handle the direction problem in KG.

\begin{table}[!htb]
\centering
\resizebox{.98\columnwidth}{!}{\begin{tabular}{c|c|c|c|c}
 \hline
  &   \multicolumn{2}{c|}{CoMPILE} & \multicolumn{2}{c}{GraIL}\\
 \hline
  & Standard & Exchanging h\&t  & Standard  & Exchanging h\&t\\
  \hline
 AUC &67.11  & 67.11 & 61.78 & 52.89\\
  AUC-PR &63.75  & 61.53 & 64.44 & 56.17\\
    \hline
 \end{tabular}}
  \caption{ \label{t6}Case Study for Asymmetric Relations in Inductive FB15k-237-v1 dataset. `Standard' denotes standard operation to sample negative triplets, and `Exchanging h\&t' denotes obtaining negative triplets by exchanging the head and tail of the test triplets.}
\end{table}%

\section{Conclusions}
We propose CoMPILE, a directed subgraph reasoning method for inductive relation prediction. We propose to use directed subgraph to infer the relation of two unseen entities which can handle asymmetric relations. Then we introduce a new message passing model that brings in the idea of enhanced edge attention and edge-node interactions to learn better node and edge embeddings and strengthen the role of relation information. Finally, we extract new inductive datasets by filtering out the triples that have no subgraph so as to evaluate each method more accurately. Experiments suggest that our CoMPILE achieves remarkable performance and outperforms state-of-the-art methods on the majority of evaluation metrics. In future  work, we aims to develop a relation-only message passing network that completely discards the nodes in the enclosing subgraph and investigate the performance of relation enclosing subgraph.

\section*{Acknowledgments}
This work was supported in part by the National Natural Science Foundation of China (62076262, 61673402, 61273270, 60802069), in part by the Natural Science Foundation of Guangdong Province (2017A030311029), in part by the Project of Guangzhou Science and Technology under Grant 201605122151511, in part by the National Key R\&D Program of China (2020YFB020000), in part by the National Natural Science Foundation of China (61772566, 62041209), in part by the Guangdong Key Field R\&D Plan (2019B020228001 and 2018B010109006),  and in part by the Guangzhou S\&T Research Plan (202007030010).

\bibliographystyle{aaai21}
\bibliography{sentiment2}

\end{document}